%% file: main.tex
\documentclass[letterpaper,times]{IONconf}

\usepackage{hyperref} 
\usepackage{url} 
\usepackage{fancyhdr}
\usepackage{courier} 
\usepackage{amsmath}
\usepackage{amsfonts}
\usepackage[euler]{textgreek}
\usepackage{bbm} 
\usepackage[]{graphicx}
\usepackage{float}
\usepackage{lipsum}
\usepackage{subcaption}
\usepackage{dsfont}
\usepackage{mathtools}
\usepackage{amssymb}
\usepackage{siunitx} 
\usepackage{bm}
\usepackage[dvipsnames]{xcolor}
\usepackage{booktabs}
\usepackage{capt-of}
\usepackage{subcaption}

\usepackage[normalem]{ulem} 

\usepackage[linesnumbered,ruled,vlined]{algorithm2e}

\SetKwInput{KwInput}{Input}                
\SetKwInOut{KwOutput}{Output}              
\SetKwInput{KwParam}{Parameters}

\usepackage[ruled,vlined,linesnumbered]{algorithm2e}
\IncMargin{0.5cm} 

\usepackage{outlines}
\usepackage{enumitem}
\setenumerate[2]{label=\alph*.}
\setenumerate[3]{label=\roman*.}

\usepackage[backend=biber,style=apa, natbib=true]{biblatex}
\include{macros}

\title{Improving GNSS Positioning using Neural Network-based Corrections}

\author{
    Ashwin~V.~Kanhere*, Shubh~Gupta*, Akshay~Shetty and Grace~Gao \\ \textit{Stanford~University}
}

\begin{document}
\date{}
\maketitle

\nnfootnote{* These authors contributed equally to this work}
\input{sections/00_biographies}
\input{sections/01_abstract}



\input{sections/02_introduction}
\input{sections/03_background}
\input{sections/04_overview}
\input{sections/05_results}
\input{sections/06_conclusion}

\input{sections/07_acknowledgements}



\printbibliography[heading=bibintoc, title={References}]

\end{document}

%% file: macros.tex
\newcommand{\edit}[1]{{\color{black}#1}}
\newcommand{\add}[1]{{\color{black}#1}}
\newcommand{\newedit}[1]{{\color{Black}#1}}

\newcommand\nnfootnote[1]{%
  \begin{NoHyper}
  \renewcommand\thefootnote{}\footnote{#1}%
  \addtocounter{footnote}{-1}%
  \end{NoHyper}
}

\newcommand{\N}{\mathbb{N}}

\newcommand{\regtext}[1]{\mathrm{\textnormal{#1}}}

\newcommand{\mc}[1]{\mathcal{#1}}

\newcommand{\norm}[1]{\left\Vert#1\right\Vert}

\newcommand{\vc}[1]{\bm{#1}}
\newcommand{\set}[1]{\bm{\mc{#1}}}
\newcommand{\transform}{\mathcal{T}}
\newcommand{\true}[1]{#1}
\newcommand{\est}[1]{\hat{#1}}

\newcommand{\samp}[1]{^{\left(#1\right)}}
\newcommand{\ind}[1]{_{#1}}

\newcommand{\pos}{\vc{p}}
\newcommand{\truepos}{\vc{\true{\pos}}}
\newcommand{\estpos}{\vc{\est{\pos}}}

\newcommand{\truecorrect}{\Delta \truepos}
\newcommand{\estcorrect}{\Delta \estpos}

\newcommand{\init}{_\regtext{init}}

\newcommand{\satpos}{\vc{P}}

\newcommand{\pseudo}{\rho}
\newcommand{\los}{\vc{1}}
\newcommand{\setlos}{\set{I}}
\newcommand{\res}{\edit{r}}
\newcommand{\setres}{\set{R}}
\newcommand{\setmeasureglob}{\set{\bar{M}}}
\newcommand{\setmeasure}{\set{M}}

\newcommand{\ecef}{_\regtext{ECEF}}
\newcommand{\ned}{_\regtext{NED}}
\newcommand{\nedtoecef}{_{\regtext{NED} \rightarrow \regtext{ECEF}}}
\newcommand{\eceftoned}{_{\regtext{ECEF} \rightarrow \regtext{NED}}}

\newcommand{\noise}{\varepsilon}

\newcommand{\batchsz}{N_b}

\newcommand{\indomain}{\bar{\set{X}}}
\newcommand{\outrange}{\bar{\set{Y}}}
\newcommand{\invalue}{\vc{X}}
\newcommand{\setinvalue}{\set{\invalue}}
\newcommand{\outvalue}{\vc{Y}}

\newcommand{\func}{\Phi}
\newcommand{\nn}{\func}
\newcommand{\varfeature}{\vc{f}}
\newcommand{\setfeature}{\set{F}}
\newcommand{\embedding}{\vc{e}}
\newcommand{\permute}[1]{\pi(#1)}
\newcommand{\nnencode}{\nn_{\regtext{encoder}}}
\newcommand{\nnaggr}{\nn_{\regtext{aggregate}}}
\newcommand{\nndecode}{\nn_{\regtext{decoder}}}

\newcommand{\funcglobal}{\tilde{\Phi}}
\newcommand{\funclocal}{\Phi}

\newcommand{\siunit}[2]{\SI[parse-numbers=false]{#1}{#2}}

\newcommand{\listicle}[1]{\bar{#1}}
\newcommand{\reslist}{\listicle{R}}
\newcommand{\loslist}{\listicle{I}}
\newcommand{\correctlist}{\Delta\listicle{P}}

\newcommand{\iono}{I}
\newcommand{\tropo}{T}
\newcommand{\isrb}{b}
\newcommand{\satbias}{B}
\newcommand{\rawpseudo}{\bar{\rho}}

%% file: sections/00_biographies.tex
\section*{biography}


\biography{Ashwin~V.~Kanhere}{is a Ph.D. candidate in the Department of Aeronautics and Astronautics at Stanford University. 
He received an M.S. in Aerospace Engineering from the University of Illinois at Urbana-Champaign in 2019 and a B.Tech in Aerospace Engineering from the Indian Institute of Technology Bombay, Mumbai in 2017. 
His research interests are in reliable navigation, with a particular interest in the fusion of Computer Vision and LiDAR applications.}

\biography{Shubh~Gupta}{is a Ph.D. candidate in the Department of Electrical Engineering at Stanford University. He received his M.S. degree in Electrical Engineering from Stanford University in 2020 and a B.Tech degree in Electrical Engineering with a minor in Computer Science and Engineering from the Indian Institute of Technology Kanpur in 2018. His research interests include safe positioning and perception for autonomous navigation.}

\biography{Akshay~Shetty}{is a postdoctoral researcher in the Department of Aeronautics and Astronautics at Stanford University. He obtained his Ph.D. in Aerospace Engineering from University of Illinois at Urbana-Champaign, and his B.Tech. in Aerospace Engineering from Indian Institute of Technology Bombay. His research interests include safe positioning and navigation of autonomous systems.}

\biography{Grace~Gao}{is an assistant professor in the Department of Aeronautics and Astronautics at Stanford University.
Before joining Stanford University, she was an assistant professor at University of Illinois at Urbana-Champaign.
She obtained her Ph.D. degree at Stanford University.
Her research is on robust and secure positioning, navigation and timing with applications to manned and unmanned aerial vehicles, robotics, and power systems.}

%% file: sections/01_abstract.tex
\section*{Abstract}
Deep Neural Networks (DNNs) are a promising tool for Global Navigation Satellite System (GNSS) positioning in the presence of multipath and non-line-of-sight errors, owing to their ability to model complex errors using data. 
However, developing a DNN for GNSS positioning presents various challenges, such as 1) poor numerical conditioning caused by large variations in measurements and position values across the globe, 2) varying number and order within the set of measurements due to changing satellite visibility, and 3) overfitting to available data.	   
In this work, we address the aforementioned challenges and propose an approach for GNSS positioning by applying DNN-based corrections to an initial position guess.
Our DNN learns to output the position correction using the set of pseudorange residuals and satellite line-of-sight vectors as inputs.
The limited variation in these input and output values improves the numerical conditioning for our DNN. 
We design our DNN architecture to combine information from the available GNSS measurements, which vary both in number and order, by leveraging recent advancements in set-based deep learning methods.
Furthermore, we present a data augmentation strategy for reducing overfitting in the DNN by randomizing the initial position guesses.
We first perform simulations and show an improvement in the initial positioning error when our DNN-based corrections are applied. After this, we demonstrate that our approach outperforms a WLS baseline on real-world data. 
Our implementation is available at~\nolinkurl{github.com/Stanford-NavLab/deep_gnss}.

%% file: sections/02_introduction.tex
\section{Introduction}\label{sec:intro}


In the last decade, deep learning has been applied in several localization applications involving complex and high-dimensional sensor inputs, such as camera images and LiDAR pointclouds~\citep{kendall_2015_posenet,wang_2017_deepvo,choy_2020_deep}. Deep learning algorithms utilize labelled data to 1) discover an effective representation, or \textit{embedding}, of the sensor inputs needed for localization, and to 2) build an approximate model, represented by a deep neural network (DNN), of the sensor input-position output relationship. Since both the embeddings and the model are learned using data, these methods have shown better performance than analytical methods when sensor inputs are affected by environmental factors, such as occlusions and dynamic obstacles~\citep{sunderhauf_limits_2018}.

Given the success of deep learning in localization using sensor inputs, it is natural to consider applying deep learning for localization using GNSS measurements. This is especially important for localization in urban and semi-urban environments, where multipath and non-line-of-sight (NLOS) effects add environment-dependent additive biases to GNSS measurements, which are challenging to model analytically. The error distributions in GNSS measurements due to these effects are often non-Gaussian, which reduces the accuracy of traditional techniques that rely on Gaussian approximations of the error~\citep{zhu_gnss_2018, reisdorf_problem_2016, wen_2020_urbanloco}. Since DNNs can learn the relationship between the measurements and corresponding positions using data, they offer a promising alternative for localization in urban and semi-urban environments.

Availability of labelled datasets containing ground truth positions is necessary for training a DNN for localization. The recent increase in public datasets containing GNSS pseudorange measurements along with the associated ground truth positions is promising for the development of deep learning algorithms for GNSS-based localization~\citep{fu_android_2020}. These datasets are collected over different driving scenarios, such as highway, urban and semi-urban, and under different operating conditions. Thus, these datasets provide a variety of input-output pairs for training the DNN.

Although labelled data with GNSS pseudorange measurements is becoming increasingly available, three main challenges must be addressed before this data can be used to train a DNN for localization: 
\begin{enumerate}
    \item \textbf{Different variations in values of the GNSS data}. Satellite positions in the earth-centered, earth-fixed (ECEF) frame of reference can take values between $\SI[parse-numbers=false]{[-20200, 20200]}{km}$ in all three axes with variations of the same magnitude.
    On the other hand, GNSS pseudorange measurements have values of around $ \SI[parse-numbers=false]{20200}{km}$ but variations on a much smaller scale, of about $\siunit{100}{m}$. 
    Similarly, GNSS receiver positions in the ECEF reference frame take values approximately between $\siunit{[-6000, 6000]}{km}$ in all three axes with variations of the same magnitude. 
    \newedit{The large difference in the ratio of meaningful variations to received values causes the optimization problem of training a DNN to be numerically ill-conditioned, resulting in}
    large changes to the DNN's parameters at each update and 
    numerical instability~\citep{goodfellow_2016_deep,mckeown_numerical_1997}.
    Furthermore, naïvely rescaling the satellite position and pseudorange measurement values risks loss of information necessary for positioning due to finite precision of floating point operations. Therefore, additional strategies for representing the satellite positions and pseudorange measurements must be considered.
    \item \textbf{Varying number and order of GNSS measurements}. Since the number of visible satellites at a measurement epoch depends on the environment, the set of measurements received at different epochs often contains different number of GNSS signals. 
    Additionally, for the same set of measurements, the output of GNSS-based localization algorithms should be independent of the order of measurements within the set. 
    However, most DNN architectures are designed for a fixed number of inputs supplied in a pre-determined order, requiring the use of specialized architectures for GNSS-based localization~\citep{lee_set_2019, skianis_rep_2020, zaheer_deep_2018}. 
    \item \textbf{Limitation in collecting vast amounts of real-world GNSS data and ground truth}. 
    Collection of large-scale GNSS datasets for deep learning is limited by the need of ground truth positions associated with the measurements, which requires sophisticated hardware. Therefore, the existing GNSS datasets with ground truth are collected at a few locations in the world and at specific times.
    These datasets are limited both in the geography and in the variety of observed pairs of GNSS measurements and positions. 
    For instance, the ECEF positions of both the receiver and the satellites captured in a dataset collected within California will not include the ECEF positions seen in a dataset collected within Australia. 
    Using such limited data in deep learning often results in DNN models that overfit on the training data and perform poorly on unseen inputs~\citep{goodfellow_2016_deep}.  
\end{enumerate}

\begin{figure}[t]
  \centering
  \includegraphics[width=0.8\linewidth]{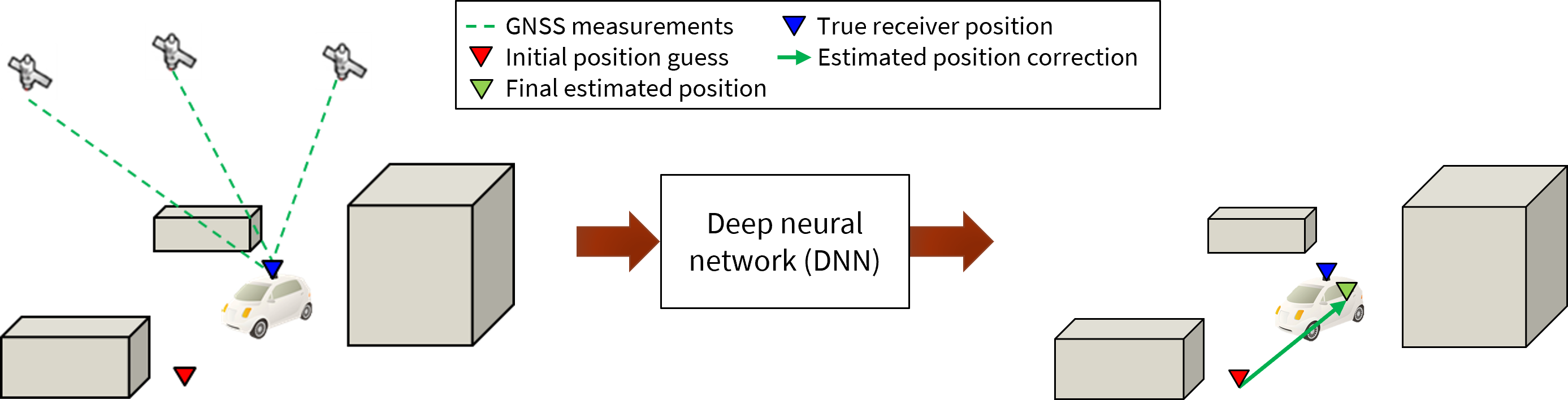} \hfill
  \caption{Our approach for applying deep learning for GNSS-based localization. Given GNSS pseudorange measurements and satellite positions, our method uses a DNN to estimate position corrections to an initial position guess.}
  \label{fig:contributions}
\end{figure}

In this work, we address these challenges and develop a deep learning algorithm for localization using GNSS pseudorange measurements. 
We propose converting the position estimation problem solved by traditional GNSS positioning algorithms into the problem of estimating position corrections to an initial position guess. 
In our approach, we use a DNN to learn a functional mapping from GNSS measurements to these position corrections, as illustrated in Fig.~\ref{fig:contributions}. This paper is based on our work in \citep*{kanhere_2021_improving}.

The main contributions of our work are:

\begin{enumerate}
    \item Design a DNN to estimate position corrections to an initial position guess. \edit{To} our knowledge, our approach is \newedit{one of }the first to use a DNN \edit{with outputs directly} in the GNSS positioning domain. 
    
    \item Use a set-based DNN architecture to handle the varying number and order of GNSS inputs at each measurement epoch.

    \item Use numerically conditioned inputs and outputs, in a local frame of reference, for the DNN. We use residuals and line-of-sight (LOS) vectors as inputs along with position correction outputs in the local North-East-Down (NED) frame of reference for numerically stable training and \edit{to encourage} global applicability of the algorithm. 
    
    \item Develop a geometry-based data augmentation strategy to prevent overfitting in the DNN and improve its generalization to new GNSS measurements. Our strategy generates new data points for training the DNN by leveraging the geometrical relationship between randomized initial position guesses, residuals, LOS vectors and position corrections.  
    
    \item Validate our proposed approach on simulations and real-world data from the Android Raw GNSS Measurements Dataset~\citep{fu_android_2020}.
\end{enumerate}

Our implementation is also publicly available at \nolinkurl{github.com/Stanford-NavLab/deep_gnss}.

The rest of this paper is organized as follows. Section~\ref{sec:related_work} discusses related work; Section~\ref{sec:set_ops} gives relevant background on set-based deep learning; Section~\ref{sec:method} provides a description of our proposed method including details of numerical conditioning of the DNN input-output values, our data augmentation strategy and the neural network architecture; Section~\ref{sec:experiments} details our experimental validation on both simulated and real-world datasets. Finally, Section~\ref{sec:conclusion} concludes this paper.

\section{Related Work}\label{sec:related_work}

Previous work has primarily used deep learning in the GNSS measurement domain for detecting faulty measurements or estimating pseudorange uncertainty. 
In~\citep{hsu_gnss_2017}, the author proposes an approach that uses a Support Vector Machine (SVM) for detection of multipath, LOS and NLOS measurements. 
The SVM is given a feature vector consisting of pseudorange residuals, pseudorange rate residuals, and signal-to-noise ratio for each measurement. The author shows that the SVM improves the NLOS, LOS and multipath detection rate compared to a \newedit{fixed}
detection threshold.
In~\citep{munin_convolutional_2019}, the authors detect multipath signals using a Convolutional Neural Network (CNN) by learning relevant visual features from the receiver correlator output for each satellite measurement.
In~\citep{zhang_prediction_2021}, the authors use a combination of Long Short-Term Memory (LSTM) and CNNs to predict satellite visibility and pseudorange uncertainty. The LSTM architecture proposed by the authors handles varying number and order of GNSS measurements in detecting multipath signals. 
However, these previous works~\citep{hsu_gnss_2017, munin_convolutional_2019, zhang_prediction_2021} focus on applying deep learning in the GNSS measurement domain and not directly in the GNSS positioning domain. 

In line with our proposed approach, several previous works have proposed estimating the pose (position and orientation) from sensor measurements by estimating and applying a correction to an initial pose guess. 
In~\citep{cattaneo_cmrnet_2019}, the authors propose a localization approach using a camera image measurement and a LiDAR map of the environment. The approach trains several DNNs to iteratively correct an initial pose guess based on a learned disparity between the camera image and an expected image constructed from the LiDAR map. 
In~\citep{peretroukhin_dpc-net_2018} the authors generate correction factors within a Factor Graph using pairwise image measurements from a camera. 
The correction factor is obtained from a DNN and represents the relative pose between the two pairs of images. Although the idea of estimating position using corrections to an initial guess has been explored in literature, it has not been applied to the problem of GNSS-based positioning using deep learning, which is the focus of this work.

%% file: sections/03_background.tex
\section{Background: Deep Learning on sets}\label{sec:set_ops}

Since the visibility of different satellites changes depending on both the location and the time of measurement, GNSS positioning output must be consistent for inputs containing different number and ordering of the measurements. For example, the position estimated using GNSS measurements from satellites numbered $1-8$ must be similar to that estimated using satellites numbered $5-10$, even if both the number of measurements and the order in which measurements from the same satellites appear are different in both cases. 
These inputs of varying size and order are commonly referred to as "set-valued inputs". 
Set-valued inputs pose unique challenges to common DNN architectures, which are designed to operate on inputs with fixed dimensions and are sensitive to the order in which different elements appear within the input~\citep{zaheer_deep_2018}.

Recently, DNN architectures that can handle set-valued inputs have been explored in literature~\citep{zaheer_deep_2018,lee_set_2019,skianis_rep_2020}. For set-valued inputs comprised of elements in domain $\indomain$ and outputs in domain $\outrange$, the objective of these DNN architectures is to learn a function $\func : 2^{\indomain} \to \outrange$, such that
\begin{gather}
    \outvalue\samp{i} = \func\left( \setinvalue\samp{i} \right), \\
    \setinvalue\samp{i} = \{\invalue\samp{i}\ind{1}, \invalue\samp{i}\ind{2}, \ldots, \invalue\samp{i}\ind{M\samp{i}}\} \quad \forall \  M\samp{i} \in \N,
\end{gather}
where $2^{\indomain}$ denotes the power set containing all combinations of elements with domain $\indomain$; $\setinvalue\samp{i}$ denotes the $i$th set-valued data instance with $\invalue\samp{i}\ind{1}, \invalue\samp{i}\ind{2}, \ldots, \invalue\samp{i}\ind{M\samp{i}} \in \indomain$; $\outvalue\samp{i} \in \outrange$ denotes the $i$th set-valued output; and $M\samp{i}$ is the number of elements in $\setinvalue\samp{i}$, which can vary across data instances.

To operate on sets, $\func$ satisfies the following two properties
\begin{enumerate}
    \item \textbf{Order invariance}. For an input $\setinvalue = \{\invalue\ind{1}, \invalue\ind{2}, \ldots, \invalue\ind{M}\}$ and its permutation $\setinvalue' = \{\invalue\ind{\permute{1}}, \invalue\ind{\permute{2}}, \ldots, \invalue\ind{\permute{M}}\}$, which has the same elements as $\setinvalue$ but with a different order defined by the operator $\permute{\cdot}$, the function output should remain the same, i.e., $f(\setinvalue) = f(\setinvalue')$.
    \item \textbf{Consistency with variable input size}. For inputs $\setinvalue = \{\invalue\ind{1}, \invalue\ind{2}, \ldots, \invalue\ind{M}\}$ and $\setinvalue' = \{\invalue\ind{1}, \invalue\ind{2}, \ldots, \invalue\ind{M'}\}$, with different number of elements ($M \neq M'$), $f$ has well-defined outputs i.e. $f(\setinvalue), f(\setinvalue') \in \mathcal{Y}$.
\end{enumerate}

DNNs equipped to handle set-valued inputs realize these properties in three main process steps: 1) generating input embeddings, 2) aggregating these embeddings and 3) processing the aggregated embeddings to produce the output~\citep{soelch_deep_2019}. 
In the following description of DNNs for set-valued inputs, we walk through these three steps for applying $\func$ to a single data instance. 
Correspondingly, we  simplify the notation from $\invalue\samp{i}$ to $\invalue$.

In the first step, an encoder network $\nnencode$, composed of feed-forward neural network layers individually processes each element $\invalue\ind{m} \ \forall \  m \in \{1, \ldots, M\}$ within the set-valued input $\setinvalue$ to obtain corresponding feature embeddings $\varfeature\ind{m}$ such that
\begin{equation}
    \varfeature\ind{m} = \nnencode(\invalue\ind{m}).
\end{equation}
For the set input, we denote this encoding process as
\begin{equation}
    \setfeature = \nnencode\left(\setinvalue\right),
\end{equation}
where $\setfeature = \lbrace \varfeature\ind{1}, \ldots \varfeature\ind{M}\rbrace$ is the set of all embeddings such that $\varfeature\ind{m} = \nnencode(\invalue\ind{m})$.

In the second step, the aggregation function combines the embeddings $\varfeature\ind{m}$ into a fixed-size aggregated embedding $\embedding$ of the inputs using an aggregation function $\nnaggr$
\begin{equation}
    \embedding = \nnaggr(\setfeature).
\end{equation}
Since the aggregation function $\nnaggr$ combines the embeddings from different input elements in the set to a fixed-size output, $\nnaggr$ can be chosen such that it is number and order invariant. 

Finally, in the third step, a decoder network $\nndecode$ composed of feed-forward neural network layers processes the embedding $\embedding$ to produce the output $\outvalue$
\begin{equation}
    \outvalue = \nndecode(\embedding).
\end{equation}

As a result of the three steps, the overall function $\func : 2^{\indomain} \to \outrange$ can be represented as
\begin{equation}
    \label{eq:set-ops-overall}
    \outvalue = \func(\setinvalue) = \nndecode\left(\nnaggr\left(\nnencode\left(\setinvalue\right)\right)\right)
\end{equation}
If the aggregation function $\nnaggr$ is chosen to be number and order invariant, the composite function $\func$ is both invariant to the ordering of the inputs and unaffected by the number of elements. A variety of aggregations $\nnaggr$ that fulfill this criteria have been studied in literature, such as sum, max-pooling, and learned aggregations~\citep{soelch_deep_2019}.

Set transformer~\citep{lee_set_2019} is a particular type of DNN architecture for set-valued inputs that uses learned aggregations to construct the fixed-size input encoding $\embedding$. In set transformers, the learned aggregations consider interactions between different set elements while combining the embeddings $\varfeature\ind{m}$. Modeling these element to element interactions has shown to perform well in tasks such as clustering, where the effective aggregation needs to be determined from the set elements themselves. Furthermore, these learned aggregations have been shown to perform well for a wide range of hyperparameters~\citep{soelch_deep_2019}. 

GNSS-based localization benefits from such considerations in modeling element-element interactions, since comparisons between different GNSS measurements aid in the detection of multipath and NLOS errors~\citep{mikhailov_identification_2012,savas_multipath_2019}. 
Additionally, the set transformer aggregation function $\nnaggr$ is number and order invariant which allows its application to set-valued inputs, such as GNSS measurements. 
Hence, we employ the set transformer within our DNN architecture to handle set-valued GNSS measurements.   

%% file: sections/04_overview.tex
\section{Proposed Method}\label{sec:method}



\begin{figure}[t]
  \centering
  \includegraphics[width=\linewidth]{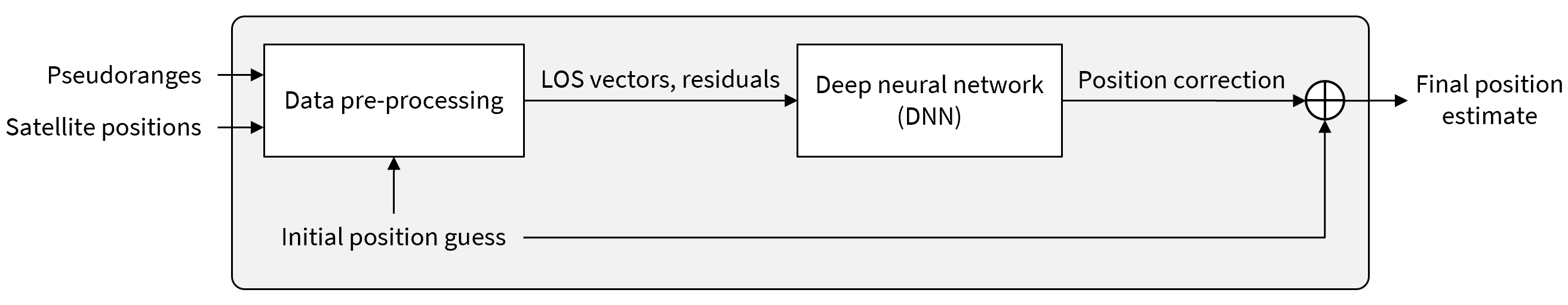} \hfill
  \caption{Overall architecture. First, we use GNSS pseudorange measurements and satellite positions to generate pairs of pseudorange residuals and LOS vectors with respect to an initial position guess as explained in Section~\ref{sec:method}1. Next, we train a DNN (Section~\ref{sec:method}2) to estimate position corrections to this initial position guess. Here while training the DNN we incorporate strategies explained in Section~\ref{sec:method}3 to improve its generalization capabilities. Finally, the position correction from our DNN is used to improve the initial position guess and obtain our final position estimate (Section~\ref{sec:method}4).}
  \label{fig:method-flowchart}
\end{figure}

In this section, we describe our approach for developing a DNN for estimating corrections, to an initial position guess, using GNSS pseudorange measurements. First, we formulate the problem of estimating position corrections with data values that are numerically well-conditioned for deep learning. Then, we describe the architecture and training process of our DNN that employs a set transformer to process the set-valued inputs derived from GNSS measurements and estimates the position correction. Next, we explain our strategies to overcome the problems of geographic sparsity of data and overfitting. Finally, we illustrate our inference procedure for a new set of GNSS measurements. Fig.~\ref{fig:method-flowchart} shows the overall architecture of our algorithm.

\input{sections/04_1_formulation}

\input{sections/04_2_architecture}

\input{sections/04_3_generalization}

\input{sections/04_4_inference}

%% file: sections/04_1_formulation.tex
\subsection{Position Correction from GNSS Measurements}\label{sec:prob}

At a measurement epoch, \edit{typical methods} estimate the position using GNSS pseudorange measurements $ \pseudo\samp{i}\ind{1}, \pseudo\samp{i}\ind{2}, \ldots, \pseudo\samp{i}\ind{M\samp{i}},$ collected from a position $\pos\samp{i}\ecef$ in the ECEF frame of reference.
Here, $i$ denotes the $i$th data instance within the dataset of GNSS measurements with $M\samp{i}$ satellites visible from $\pos\samp{i}\ecef$. 
The position $\satpos\samp{i}\ind{m}$ of the satellite associated with each $m$th measurement is available from the ephemerides broadcast by the International GPS Service (IGS) ground stations. 
\edit{For inputs} to typical GNSS localization algorithms, consider the set $\setmeasureglob\samp{i}$ of pairs of pseudoranges with corresponding satellite positions, defined as
\begin{equation}
    \setmeasureglob\samp{i} = \left\lbrace \left(\pseudo\samp{i}\ind{1}, \satpos\samp{i}\ind{1}\right), \left(\pseudo\samp{i}\ind{2}, \satpos\samp{i}\ind{2}\right),\ldots \left(\pseudo\samp{i}\ind{M\samp{i}}, \satpos\samp{i}\ind{\ind{M\samp{i}}}\right)\right\rbrace.
\end{equation}
\edit{In a machine learning setting, the} objective is to learn a function $\funcglobal$ that outputs an estimate $\estpos\samp{i}\ecef$ of the position  $\truepos\samp{i}\ecef$ using the input $\setmeasureglob\samp{i}$, for the $i$th data instance, where $N$ is total number of \edit{instances} in the dataset,
\begin{equation}
    \estpos\samp{i}\ecef = \funcglobal\left(\setmeasureglob\samp{i}\right) \quad \forall \ i\in \{1,\ldots, N\}. \label{eq:form_1}
\end{equation}
As previously \edit{mentioned} in Sec. \ref{sec:intro}, Eq.~\eqref{eq:form_1} has poor numerical conditioning  for deep learning due to large variations in the input and output values.
Therefore, we rewrite Eq.~\eqref{eq:form_1} \edit{as the addition of an estimated position correction $\estcorrect\samp{i}\ecef$ to an initial position guess $\estpos\init\samp{i}$} 

\begin{equation}
    \estpos\samp{i}\ecef = \edit{\estpos}\samp{i}\init + \estcorrect\samp{i}\ecef.
\end{equation}

\edit{In this new setting, } \edit{the} objective is to learn a function $\funcglobal$ that outputs an estimate of the position correction $\estcorrect\samp{i}\ecef$ using the input $\setmeasureglob\samp{i}$, such that
\begin{equation}
    \estcorrect\samp{i}\ecef = \funcglobal(\setmeasureglob\samp{i}, \add{\estpos\samp{i}\init}) \quad \forall \ i\in \{1,\ldots, N\}. \label{eq:form_2}
\end{equation}

If $\edit{\estpos}\samp{i}\init$ is selected in the vicinity of the true position $\truepos\samp{i}\ecef$, the different values \edit{of} the true position correction $\truecorrect\samp{i}\ecef = \truepos\samp{i}\ecef - \edit{\estpos}\samp{i}\init$ are restricted to a small range, as opposed to the large variations in $\truepos\samp{i}\ecef$, \edit{resulting in better conditioning of the output for function $\funcglobal$}. 
 
To improve the conditioning of the input variables, we consider the pseudorange model~\citep{morton_position_2021}
 \begin{gather}
     \pseudo\samp{i}\ind{m} = \norm{\satpos\samp{i}\ind{m} - (\edit{\estpos}\samp{i}\init + \Delta \truepos\samp{i}\ecef)} + \noise\samp{i}\ind{m}, \label{eq:prange_model}
 \end{gather}
where $\noise\samp{i}\ind{m}$ denotes the error in the measurement due to both factors that can be modelled, such as satellite and receiver clock drift and atmospheric effects, as well as factors that are difficult to model analytically, such as effects of multipath and NLOS signals. 

Assuming $\edit{\norm{\truecorrect\samp{i}\ecef}} <<  \norm{\satpos\samp{i}\ind{m} - \edit{\estpos}\samp{i}\init}$, we linearize the expression in Eq. \eqref{eq:prange_model} about $\edit{\estpos}\samp{i}\init$
\begin{align}
    \pseudo\samp{i}\ind{m} - \norm{\satpos\samp{i}\ind{m} - \edit{\estpos}\samp{i}\init} &\approx \left.\nabla \norm{\satpos\samp{i}\ind{m} - (\edit{\estpos}\samp{i}\init + \truecorrect)}\right|_{\truecorrect=0}  \cdot \Delta \truepos\samp{i}\ecef + \noise\samp{i}\ind{m}, \\
    \implies \res\samp{i}\ind{m} &\approx -\los\samp{i}\ind{m}  \cdot \Delta \truepos\samp{i}\ecef + \noise\samp{i}\ind{m},
    \label{eq:lin_model}
 \end{align}
where $\res\samp{i}\ind{m}$ denotes the pseudorange residual \edit{(difference between $\pseudo\samp{i}\ind{m}$ and expected pseudorange at $\estpos\init\samp{i}$)} for the $m$th satellite and $\los\samp{i}\ind{m}$ denotes the LOS vector for the $m$th satellite from $\edit{\estpos}\init$, both for the $i$th data instance. 
\edit{Thus, given an initial guess $\estpos\init\samp{i}$, we convert the set of received measurements $\setmeasureglob$ into a set containing pairs of residuals and corresponding LOS vectors, defined as}
\begin{equation}
    \setmeasure\samp{i} = \left\lbrace \left(\res\samp{i}\ind{1}, \los\samp{i}\ind{1}\right), \left(\res\samp{i}\ind{2}, \los\samp{i}\ind{2}\right),\ldots \left(\res\samp{i}\ind{M\samp{i}}, \los\samp{i}\ind{\ind{M\samp{i}}}\right)\right\rbrace,
\end{equation}
which is equivalently represented as
\begin{equation}
    \setmeasure\samp{i} = \left( \setres\samp{i}, \setlos\samp{i} \right)\label{eq:input_sets} ,
\end{equation}
where $\setres\samp{i} = \left\lbrace \res\samp{i}\ind{1}, \ldots, \res\samp{i}\ind{M} \right\rbrace$ and $\setlos\samp{i} = \left\lbrace \los\samp{i}\ind{1}, \ldots, \los\samp{i}\ind{M} \right\rbrace$.

\edit{Assuming $\estpos\init\samp{i}$ is in the vicinity of the true position, the residuals $\res\samp{i}\ind{m}$ typically vary in the range of tens of meters while the LOS vectors $\los\samp{i}\ind{m}$ are unit constrained.}
\edit{Hence, the measurement set $\setmeasure$, defined in Eq.~\eqref{eq:input_sets} is a better conditioned input to the DNN than the received measurement set $\setmeasureglob$.}


As a result of the \edit{input and output} conditioning, the \edit{effective} objective of the DNN \edit{in our approach} is to learn a functional mapping $\funclocal$ such that
\begin{equation}
    \estcorrect\samp{i}\ecef = \funclocal(\setmeasure\samp{i}) \quad \forall \ i\in \{1,\ldots, N\}. \label{eq:form_feat}
\end{equation}

\add{To summarize, the input to the network is a set $\setmeasure\samp{i}$ of residuals with corresponding LOS vectors and the output of the network is the position correction $\estcorrect\samp{i}$ at the $i$th sample of the dataset.}

%% file: sections/04_2_architecture.tex
\subsection{DNN for Estimating Position Corrections}\label{sec:arch}
To obtain the estimated position corrections $\estcorrect\samp{i}\ecef$ from the conditioned set-valued inputs $\setmeasure\samp{i}$ using Eq. \ref{eq:form_feat}, we develop a neural network based on the set transformer~\citep{lee_set_2019} architecture discussed in Section~\ref{sec:set_ops}

\begin{figure}[t]
  \centering
  \includegraphics[width=\linewidth]{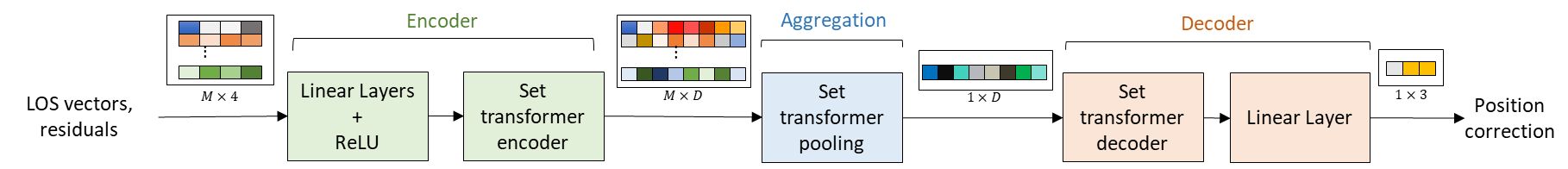} \hfill
  \caption{
  \newedit{Our DNN architecture.} The network consists of three steps: the encoder, the aggregation and the decoder. First, the LOS vectors and the pseudorange residuals from $M$ satellites are concatenated into a $M \times 4$ matrix. The encoder, comprising of fully connected layers and the set transformer encoder block~\citep{lee_set_2019}, then produces a $1 \times D$ embedding for each satellite, outputting a $M \times D$ matrix. \edit{Throughout the network, $D$ is a hyperparameter for the dimension of the latent space that inputs are projected to and feature embeddings are computed in}.  Next, the aggregation function, or the set transformer pooling block~\citep{lee_set_2019}, combines these embeddings into a fixed-size embedding and outputs a $1 \times D$ array. Finally, the decoder outputs the 3-dimensional position corrections.}
  \label{fig:DNN-architecture}
\end{figure}


Our DNN architecture comprises of four components which we train together to estimate the position corrections from input set $\setmeasure$ of residuals and LOS vectors. 
First, as a part of the encoder $\nnencode$, a fully connected network, \newedit{with ReLU activation functions,} generates a high-dimensional embedding of each input\edit{, comprising of a} residual $\res$ from $\setres$ and the associated LOS vector $\los$ from $\setlos$. 
\edit{Each embedding is a $D$-dimensional vector and is an instance of a measurement in the latent space.}
\edit{Here $D$ is a hyper-parameter of the network architecture and can be different} 
\newedit{ after encoding or after aggregation}
\edit{in the network.}
\edit{In this work, we choose $D=64$ as the hyperparameter throughout the network.}
Then, a set transformer encoder based on the set transformer encoder block~\citep{lee_set_2019} further refines the embeddings by modeling interactions between different set elements. Next, a network for learned aggregation $\nnaggr$, based on set transformer pooling block~\citep{lee_set_2019}, determines the influence of each set element on the position correction output and combines the embeddings based on these influences. Finally, a set transformer decoder network $\nndecode$, composed of \newedit{a set transformer decoder and a linear layer}
~\citep{goodfellow_2016_deep}, processes the aggregated embedding to determine the position correction output $\estcorrect\ecef$. 
\add{Section~\ref{sec:set_ops} briefly explains the set transformer encoder, aggregation and decoder blocks.}
Fig.~\ref{fig:DNN-architecture} depicts the DNN architecture for our proposed approach. 

We train the DNN by minimizing the mean-squared-error (MSE) between a batch of the estimated and the true corrections as the loss function

\begin{gather}
    \text{MSE} = \frac{1}{\batchsz}\sum^{\batchsz}_{j=1} \norm{\truecorrect\samp{j}\ecef - \estcorrect\samp{j}\ecef}^2, 
\end{gather}
where $\batchsz$ is the batch size. 

%% file: sections/04_3_generalization.tex
\subsection{Strategies For DNN Generalization}\label{sec:generalization}
While a DNN trained from Eq. \ref{eq:form_feat} has access to well-conditioned inputs and outputs, its ability to generalize to new data instances is limited by 1) the geographic sparsity of the data, and 2) variations in inputs and outputs encountered during training. 
In this subsection, we present strategies to overcome these limitations and improve the generalization capability of the DNN.

\subsubsection{Overcoming Geographic Sparsity by Change of Reference Frame}\label{sec:geo_sparse}
Geographic sparsity arises because the dataset is collected at fixed regions on the globe. 
The satellite LOS vectors and position corrections in the data collected in one part of the world may significantly differ from that in data from another part, resulting in measurements from some regions being disproportionately represented in the dataset.
This disproportionality increases the difficulty in training the DNN to accurately estimate corrections for positions all around the globe, since certain input-output relations might be missing from the dataset.

To incentivize the DNN to generalize to inputs from across the globe, we make the input-output behavior independent of the location the data was collected in. 
We achieve this by changing the frame of reference of the inputs and outputs from the global ECEF frame to the local North-East-Down (NED) frame about $\truepos\samp{i}\init$. 
In the inputs, the residuals remain the same while the LOS vectors change. 
We represent this transformation as
\begin{equation}
    \transform\samp{i}\eceftoned\left(\setmeasure\samp{i}\right) = \left( \setres\samp{i}, \transform\samp{i}\eceftoned\left(\setlos\samp{i}\right) \right),
\end{equation}
where $\setres\samp{i}$ and $\setlos\samp{i}$ are defined the same as in Eq.~\eqref{eq:input_sets} and $\transform\samp{i}\eceftoned$ denotes the transformation from the ECEF frame to the NED frame.

The neural network estimates $\estcorrect\samp{i}\ecef$ by estimating  $\estcorrect\samp{i}\ned$ which gets modified according to the transformation
\begin{align}
    \estcorrect\samp{i}\ecef &= \transform\samp{i}\nedtoecef\left(\estcorrect\samp{i}\ned\right), \\
    \estcorrect\samp{i}\ned &= \funclocal\left(\transform\samp{i}\eceftoned\left(\setmeasure\samp{i}\right)\right) \quad \forall \ i\in \{1,\ldots, N\}, \label{eq:ned}
\end{align}
where $\transform\samp{i}\nedtoecef$ denotes the transformation from the NED frame to the ECEF frame. 

\add{The above mentioned transformation changes the inputs and outputs to a local frame of reference. 
The modified inputs and outputs make our proposed architecture amenable to measurements from parts of the world that were not seen during the training process, as measurements are similar to the network's training data after transformation.}


\subsubsection{Increasing Data Variation using Geometry-based Data Augmentation}\label{sec:var_sparse}
\begin{algorithm}[b]
\DontPrintSemicolon
  \KwIn{Set $\setmeasureglob$ of paired pseudorange measurements and satellite positions and ground truth position $\pos\ecef$}
  \KwParam{Number of augmented data points $K$ and vector-valued initialization range $\vc{\eta}$}
  \KwOut{A list of residuals $\reslist$, LOS vectors $\loslist$, and position corrections $\correctlist$}
  $\reslist \gets [\ ]  , \loslist \gets [\ ], \correctlist \gets [\ ] $ \\
  \For{$k \gets 1$ \KwTo $K$}{
        Sample $\pos\init$ uniformly from $[\pos\ecef - \vc{\eta}, \pos\ecef + \vc{\eta}]$\\
        Generate $\setres, \setlos, \truecorrect\ecef$ from  $\setmeasureglob$, and $\pos\init$ using Eq.~\eqref{eq:lin_model}\\
        Assign values $\reslist[k] \gets \setres, \loslist[k] \gets \setlos, \correctlist[k] \gets \truecorrect\ecef$
    }
    \Return{$\reslist, \loslist, \correctlist$}
\caption{Geometry-based data augmentation}
\label{alg:data_augmentation}
\end{algorithm}
Using limited data to train a DNN with several parameters often leads to overfitting, where the DNN memorizes input-output pairs specific to the training dataset~\citep{goodfellow_2016_deep}. 
Data augmentation is a commonly used technique to reduce overfitting which introduces new data points to the DNN during training by transforming existing training samples based on the problem context. 

We introduce a geometry-based data augmentation strategy for training a DNN to estimate position corrections from pseudorange measurement residuals and LOS vectors. Alg.~\ref{alg:data_augmentation} illustrates the process for generating new data points from a data instance. 
Our augmentation strategy leverages the geometric aspect of GNSS-based positioning by changing the value of the initial position guess $\edit{\estpos}\samp{i}\init$ each training epoch to generate new residuals $\setres\samp{i}$, LOS vectors $\setlos\samp{i}$ and corrections $\truecorrect\samp{i}\ecef$ via Eq. (\ref{eq:lin_model}).
\add{New initial position guesses are generated by adding zero-mean uniformly distributed noise to the ground truth position $\truepos\samp{i}$.}
\add{As a result, new samples are generated without any correlation
, thus regularizing the training process and allowing the network to better learn the input-output mapping relationship. }
\add{Finally, the network sees new samples in every training epoch, which prevents it from overfitting on the training data.}


%% file: sections/04_4_inference.tex
\subsection{Inference}\label{sec:inference}
In this section, we illustrate our process to use the trained DNN for estimating the position $\estpos\ecef$ from new GNSS pseudorange measurements and the corresponding satellite positions, represented by the set $\setmeasureglob$.

First, we obtain an initial position guess $\truepos\init$, from a traditional localization algorithm  or using prior knowledge, that we assume is in the vicinity of the true position $\truepos\ecef$. 
Then, we use Eq.~\eqref{eq:form_feat} to determine the input set $\setmeasure$ that comprises of pseudorange residuals $\setres$ and corresponding LOS vectors $\setlos$ in the NED reference frame with respect to $\truepos\init$.
Using the set $\setmeasure$ as an input to the DNN, we evaluate the position correction in the NED frame $\estcorrect\ned$ and convert it to the position correction in the ECEF frame $\estcorrect\ecef$. 
Finally, we add the correction $\estcorrect\ecef$ to $\pos\init$ to obtain the position estimate $\estpos\ecef$ using
\begin{equation}
    \estpos\ecef = \truepos\init + \estcorrect\ecef.
\end{equation}

%% file: sections/05_results.tex
\section{Experiments}\label{sec:experiments}

We validate our approach using a simulated dataset and real-world measurements from the Android Raw GNSS Measurements Dataset~\citep{fu_android_2020}.
We use simulations to verify the performance of our network in a setting with controlled measurement errors and access to precise grouth truth information.
In the validation on real-world data, we compare the accuracy of our proposed approach to that of \edit{weighted least squares (WLS)}~\citep{morton_position_2021}, which is an equivalent traditional localization algorithm \edit{and} serves as a baseline comparison. 
\edit{In experiments on both data types}, \add{we used the same network architecture, optimizer parameters, data generalization method, and other experimental hyperparameters.}
\add{These parameters are described in Section~\ref{sec:experiments}~\ref{sec:exp_params}, followed by experimental evaluation on the simulated dataset in Section~\ref{sec:experiments}~\ref{sec:sim_trajectories} and evaluation on the Android Raw GNSS Measurements Dataset in Section~\ref{sec:experiments}.~\ref{sec:google_dataset}}


\input{sections/05_0_experiments_setup}

\input{sections/05_1_experiments_sim}


\input{sections/05_2_experiments_real}


%% file: sections/05_0_experiments_setup.tex
\subsection{Experimental Parameters}
\label{sec:exp_params}
\add{In our experiments, a fully trained network occupies $611\si{kB}$ on disk for $151,107$ parameters.
We use an instance of the network described in Section~\ref{sec:method}~\ref{sec:arch} where the inputs (residuals and LOS vectors) are projected into a latent space of dimension $D=64$ by a linear layer, followed by a ReLU activation~\citep{goodfellow_2016_deep}.
In our implementation, we chose $D=64$ as the dimension of the latent spaces in which all projected and embedded features exist.

The projected features are then encoded by two Transformer Encoder layers~\citep{vaswani_attention_2017} that operate on the features sequentially.
The encoded features are pooled using a pooling attention module~\citep{lee_set_2019}, which is followed by two sequential Transformer Decoder layers and a linear layer to output the 3D position correction.}
\add{We do not use batch normalization or dropout techniques at any point in the network architecture.}

\add{Our experiments are performed with data batches of 64 samples and the network is trained for 200 epochs.
The DNN parameters are optimized using Adam~\citep{kingma_adam_2014} with a learning rate $\alpha= 3\times 10^{-4}$ and moving window average $\beta_1=0.9$, and $\beta_2=0.99$.}

\edit{At each training and testing epoch, we generate} the initial position guess $\edit{\estpos}\init$ 
by uniformly sampling from the interval $[\pos\ecef - \vc{\eta}, \pos\ecef + \vc{\eta}]$, where $\vc{\eta} = \eta \cdot [1, 1, 1]^\top$ is the vector-valued initialization range with a magnitude $\eta$ that is the same along each direction.
\add{We use initial position guesses with randomly sampled noise added to the true position, in all our experiments, except without data augmentation, for training the network and validating/testing the trained network.}
\edit{The default value in the experimental validations is $\eta=15\si{m}$, which is changed when studying the effect of different $\eta$ values on the final position estimate.}

\add{Additionally, when evaluating the effectiveness of our data augmentation method, we compare our approach \newedit {to}
a baseline without data augmentation.
In the network without data augmentation, we use a fixed trajectory uniformly sampled  from the interval $[\pos\ecef - \vc{\eta}, \pos\ecef + \vc{\eta}]$.
Here, fixed implies that the samples are drawn once to generate the training and validation datasets and are not changed at any epoch during training.}

%% file: sections/05_1_experiments_sim.tex
\subsection{Simulated Dataset}\label{sec:sim_trajectories}

We 
\edit{create} the simulated dataset by 1) generating \edit{smooth} horizontal trajectories in the NED frame of reference, 2) converting the simulated trajectories to the ECEF frame of reference, and 3) simulating open sky GNSS measurements for each point along the trajectory.


\newedit{We simulate trajectories to imitate real-world datasets, \newedit{like the Android Raw GNSS Measurements Dataset~\citep{fu_android_2020},} that are often confined to a limited geographical region and contain samples along vehicle trajectories.
We simulate these trajectories based on the approach proposed by Mueller \textit{et al.} in~\citep{mueller_computationally_2015}.}
\add{
Note that our network performs snapshot position estimation, i.e., the correlation between samples in the trajectory has no impact on our experimental results.}

To \edit{generate} measurements \edit{for samples from the simulated trajectories}, we use the standard pseudorange model~\citep{morton_position_2021} with the true position and clock states for each instance of data in the converted trajectories $\truepos\samp{i}\ecef$. We do not consider any atmospheric effects or satellite clock biases in simulating the pseudorange measurements. Set $\setmeasureglob\samp{i}$ represents the pairs of simulated pseudorange measurements and the corresponding satellite positions.

For each data instance, measurements are only simulated for satellites that are visible from $\truepos\ecef\samp{i}$, determined using an elevation mask of $5^\circ$.
Because we use an elevation mask to simulate the measurements, the number of measurements at each instance $M\samp{i}$ varies between 8-10 in our dataset.
Additionally, we impose no constraints on the order of the simulated measurements.

We next describe experiments that utilize the simulated data to verify the \edit{validity} of our approach. Additionally, we investigate the sensitivity of the DNN performance to the choice of measurement errors and the initialization range magnitude $\eta$.

\subsubsection{Verifying performance under different measurement errors}\label{sec:sim_bias}

\begin{table}[b]
    \centering
    \caption{\edit{Mean absolute positioning error} along the north, east and down directions in the position estimated using our approach across different types of error (Gaussian erro and Gaussian + bias error) in the GNSS pseudorange measurements. In both scenarios, our approach reduces the positioning error over the baseline with random initialization by more than half the value.}
    \begin{tabular}{llll}
    \toprule
    \textbf{Scenario}     & \textbf{North (m)} & \textbf{East (m)} & \textbf{Down (m)} \\ \midrule
    Initialization              & $7.5 \pm 5.0$            & $7.5 \pm 5.0$            & $7.5 \pm 5.0$            \\
    Gaussian error        & $2.6 \pm 2.0$           & $2.4 \pm 1.8$            & $2.2 \pm 1.6$           \\
    Gaussian + bias error & $2.8 \pm 2.1$           & $2.6 \pm 2.0$            & $2.4 \pm 1.8$ \\ \bottomrule \\
    \end{tabular}
    \label{tab:sim_bias}
\end{table}

We verify the positioning performance of our DNN in our approach across two scenarios with different error profiles in the pseudorange measurements. 

In the first scenario, simulated pseudoranges contain stochastic noise terms that follow a zero-mean Gaussian distribution with $\siunit{6}{m}$ standard deviation. 
In the second scenario, we add bias errors along with the zero-mean Gaussian errors in the measurements. The bias errors are sampled from the interval $\siunit{[50, 200]}{m}$ and are added to pseudoranges picked at random to mimic the effect of multipath and NLOS signals. The number of biased measurements at a time is sampled from a Poisson distribution with rate $1$.
In both scenarios, we use $\eta = \siunit{15}{m}$ for generating the initial position guess $\edit{\estpos}\init$. 
Because the DNN is not restricted by a prior measurement model, we hypothesize that the positioning error for the DNN should be unaffected by the noise scenarios, as long as the DNN encounters the same noise scenario during the training process.

To verify this hypothesis, we evaluate the \edit{mean absolute positioning error} along the north, east and down directions respectively. 
For both scenarios, the positions estimated by applying corrections from our trained DNN exhibit positioning errors that are less than half the initial value, verifying that our proposed approach is effective in learning a function for positioning using GNSS measurements.
These results are summarized in Table~\ref{tab:sim_bias}. 

\subsubsection{Comparing performance across different initial positions}

\begin{figure}[t]
    \centering
    \includegraphics[width=0.6\textwidth]{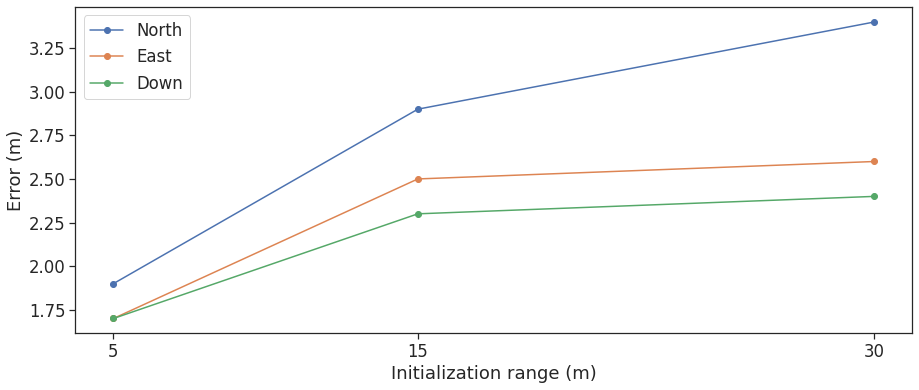}
    \captionof{figure}{Sensitivity analysis over various initialization ranges along the north, east and down directions. The \edit{mean absolute error (MAE)} in DNN-based position corrections increases when the initialization range is increased.}
    \label{fig:sim_init}
\end{figure}

Since the magnitude of the initialization range $\eta$ determines the maximum initial positioning error, we expect it to have a significant effect on the positioning performance of the DNN.
To investigate this, we evaluate the sensitivity of our approach to different choices of $\eta$ for a scenario with zero-mean Gaussian errors in pseudorange measurements. We consider three different values of $\eta \in \{\siunit{5}{m}, \siunit{15}{m}, \siunit{30}{m}\}$ for training the DNN and compare the positioning performance of the resultant DNN, the results of which are shown in Fig.~\ref{fig:sim_init}.

We observed that the positioning error along each of the north, east and down directions increases as we increase the value of $\eta$.
However, this increase isn't linear and the difference between the positioning errors for $\eta = \siunit{15}{m}$ and $\eta = \siunit{30}{m}$ shows a less than linear growth.
This indicates that while the positioning error of the DNN does depend on the magnitude of the initialization range $\eta$, the impact of $\eta$ reduces as its magnitude increases. 


\add{We attribute the increase in the mean absolute error (MAE) on increasing the initialization range $\eta$ to primarily two factors: 1) The network learns the maximum possible corrections based on the magnitude of the maximum error it sees in the training dataset. 
As a result, outputs for smaller values of $\eta$ are restricted to smaller ranges, resulting in a smaller MAE.
2) On increasing $\eta$, the network must generalize to a larger set of possible inputs, which increases the overall error in the position estimate.}


%% file: sections/05_2_experiments_real.tex
\subsection{Android Raw GNSS Measurements Dataset}\label{sec:google_dataset}


The Android Raw GNSS Measurements Dataset~\citep{fu_android_2020} consists of GNSS measurements collected using Android phones from multiple driving trajectories executed in the San Francisco Bay Area. 
This dataset has two components: 1) a training component and 2) a testing component. 
The training component is accompanied by high accuracy position estimates, collected using a \edit{NovAtel} SPAN system, that we use as the ground truth position in our approach.
Due to this \edit{availability} of ground truth positions, we restrict ourselves to the training component because the ground truth provides a reference to both train and evaluate the DNN. 
Henceforth, we refer to this training component as the dataset for evaluating our approach.
The GNSS measurements in each trajectory, referred to as a trace, include raw pseudoranges, atmospheric biases, satellite clock biases and satellite positions, from at least two Android phones.
\add{These measurements, including satellite positions, atmospheric biases and satellite clock biases, are computed and provided in `Derived' files in the dataset.
We use these quantities without any modification or additional computations.}
We treat each unique phone-trace combination as an independent trajectory while validating our approach.

To create the set $\setmeasure\samp{i}$, for each data instance that is input to the DNN, we use measurements corresponding to GPS L1 signals and process the raw pseudoranges to remove errors that can be modeled.
The corrected pseudorange $\pseudo\samp{i}\ind{m}$ is obtained from values present in the measurement dataset by
\begin{equation}
    \pseudo\samp{i}\ind{m} = \rawpseudo\samp{i}\ind{m} + \satbias\samp{i}\ind{m} -\isrb\samp{i}\ind{m} - \iono\samp{i}\ind{m} -\tropo\samp{i}\ind{m},
    \label{eq:android_pseudo}
\end{equation}
where $\rawpseudo\samp{i}\ind{m}$ represents the raw pseudorange, $\satbias\samp{i}\ind{m}$ the satellite clock bias, $\isrb\samp{i}\ind{m}$ the inter-signal-ranging-bias, $\iono\samp{i}\ind{m}$ the modeled delay due to ionospheric effects and $\tropo\samp{i}\ind{m}$ represents the modeled delay due to tropospheric effects.
This process is repeated for all measurements $m \in \lbrace1, \ldots, M\samp{i} \rbrace$ in all data instances $i \in \lbrace 1, \ldots, N \rbrace$, where $M\samp{i}$ is the number of measurements in the $i$th data instance and there are $N$ data instances in the entire dataset.

\edit{In our experimental evaluation on the Android dataset, we split the dataset into three independent parts: 1) a training split ($\approx 75\%$ of the dataset), 2) a validation split ($\approx 10\%$ of the dataset), and a 3) testing split ($\approx 15\%$ of the dataset). }

\begin{figure}
    \centering
    \includegraphics[width=0.5\linewidth]{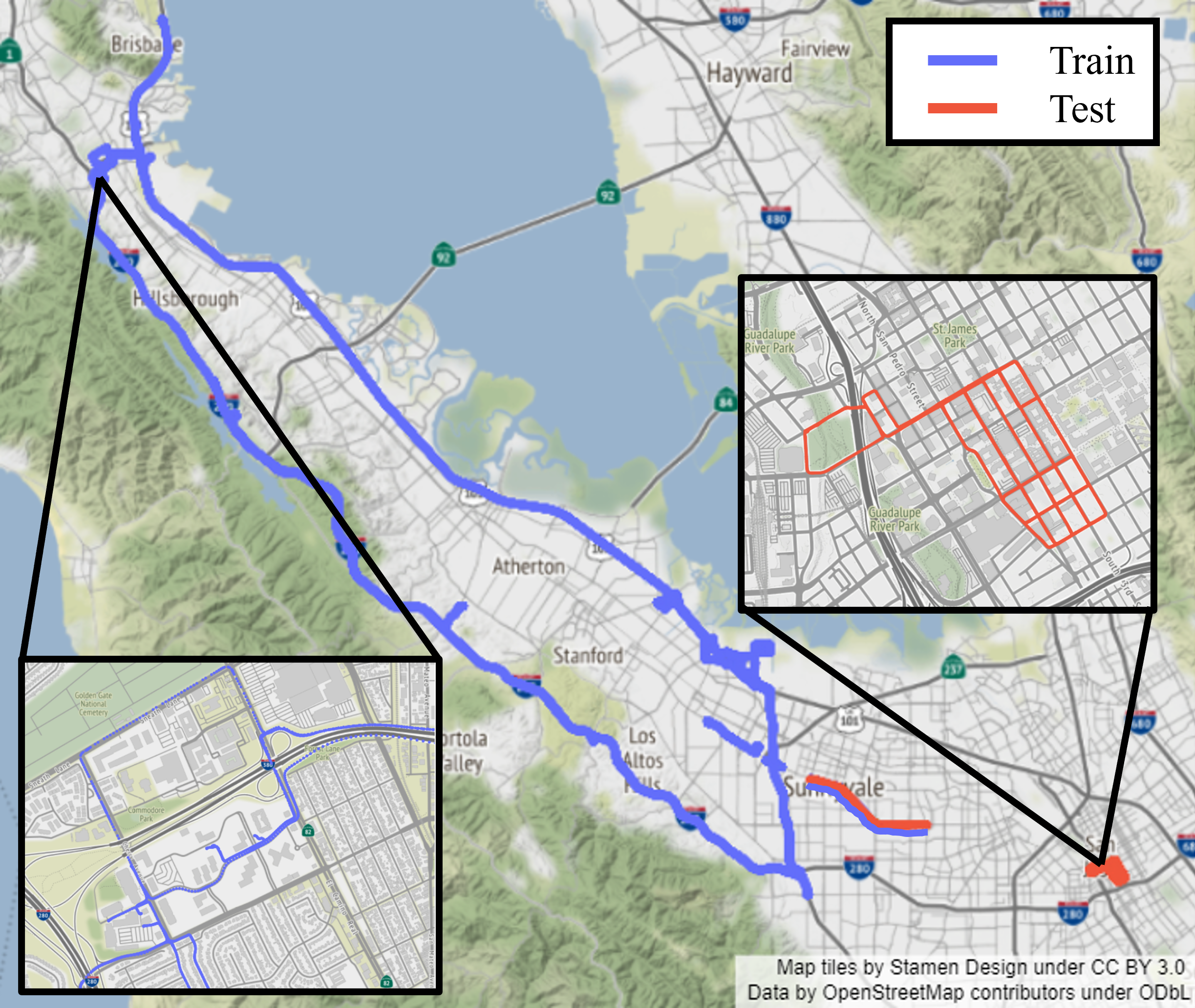}
    \caption{\add{Traces from the Android Raw GNSS Measurements dataset. Only traces with ground truth position estimates available were used for our experimental evaluation. Traces belonging to the training/validation dataset are shown in blue, while traces belonging to the testing dataset are shown in red.}}
    \label{fig:dataset}
\end{figure}

\add{The first split divides the dataset into two parts: one for training/validation and another for testing. 
This division is performed on the trace level and the training/validation and testing dataset contain different traces, with all corresponding Android measurements from a particular trace associated with either the training/validation or testing dataset.
The split between the training/validation and testing datasets is fixed and the same for all experiments in this work.
The traces belonging to each dataset are plotted in Fig.~\ref{fig:dataset}.
The additional split between the training and validation datasets is performed by randomly selecting a ratio of samples from the training/validation traces and using them to validate the network.
Each split between the training and validation dataset is stochastic and changes from experiment to experiment.

As a result of the dataset split, the training dataset has $93195$ samples, the validation dataset has $10355$ samples, and the testing dataset has $16568$ samples.
}


\subsubsection{Performance Evaluation}
\begin{figure}[t]
\centering
\begin{subfigure}{.5\textwidth}
  \centering
  \includegraphics[width=.85\linewidth]{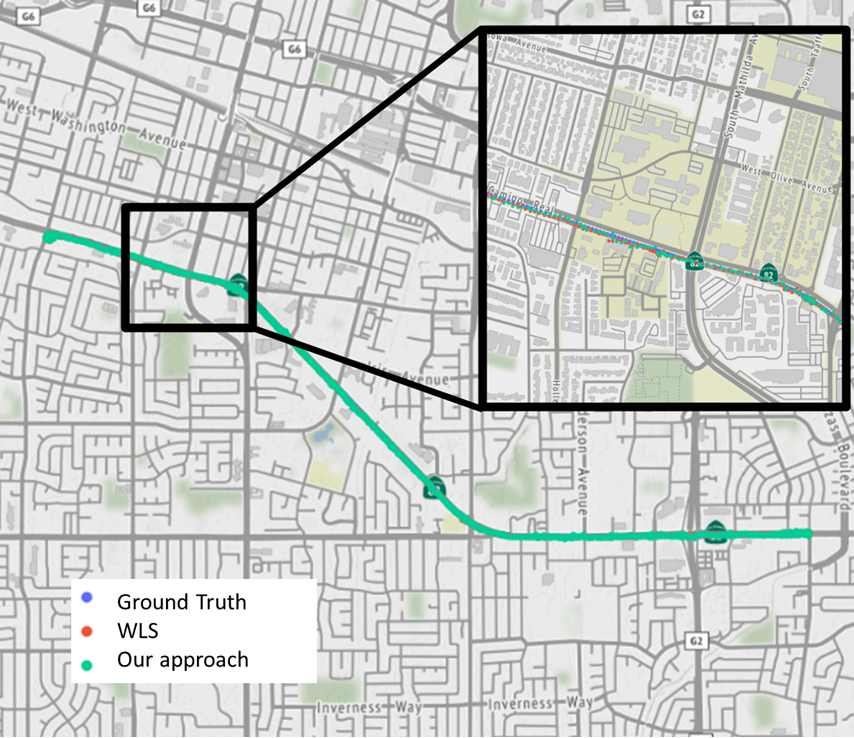}
  \caption{Semi-urban environment}
\end{subfigure}%
\begin{subfigure}{.5\textwidth}
  \centering
  \includegraphics[width=.85\linewidth]{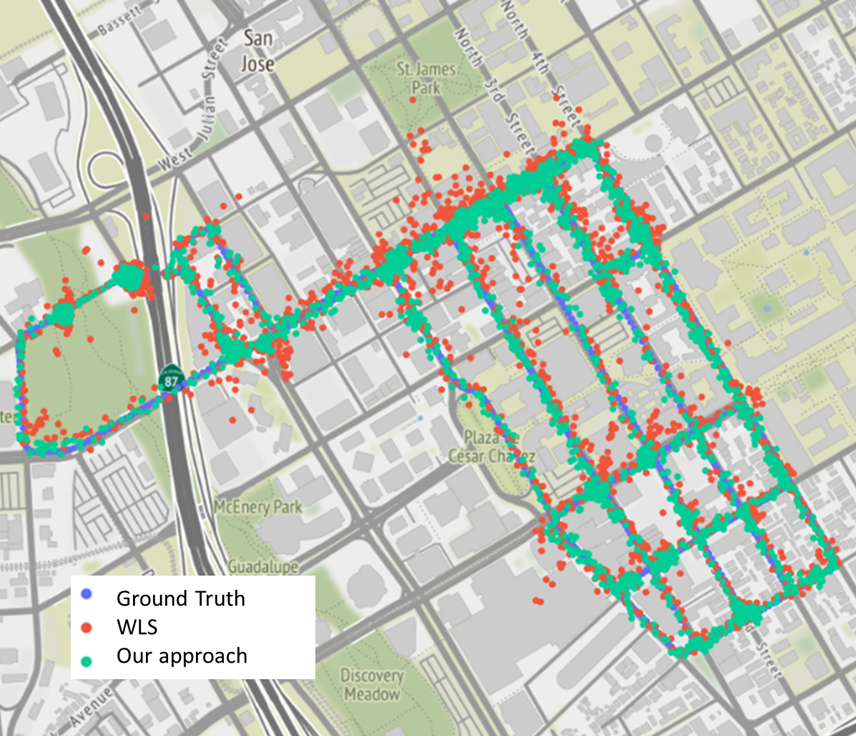}
  \caption{Urban environment}
\end{subfigure}
\caption{Example of localized trajectories on the Android Raw GNSS Measurements Dataset~\citep{fu_android_2020} for a) semi-urban and b) urban environment conditions. We visualize the positions estimated using our approach for $\eta=\siunit{15}{m}$ and WLS along with the ground truth trajectory. The trajectory estimated using our approach is visually similar to WLS in the semi-urban environment and appears closer to the ground truth than WLS in the urban environment. 
}
\label{fig:visual_realdata}
\end{figure}


We use the training split to train the DNN while the validation split is used to evaluate the DNN during training and ensure that it is learning successfully. 
We use the testing split to evaluate the performance of \edit{different variations of our approach} and compare it to the WLS baseline.

\add{The WLS baseline position estimates are generated using the open-source goGPS implementation~\citep{herrera_2016_gogps}.
goGPS internally corrects pseudoranges by removing estimated atmospheric delays, satellite clock biases and other modelled biases.
An elevation mask of $10^\circ$ was applied on the received measurements and the remaining measurements were weighed using the default elevation-based weights from goGPS.
The WLS output contained a 3D position estimate along with a clock bias estimate, of which we compare only the positions to those obtained by our proposed architecture.}



\begin{figure}
    \centering
    \includegraphics[width=0.8\linewidth]{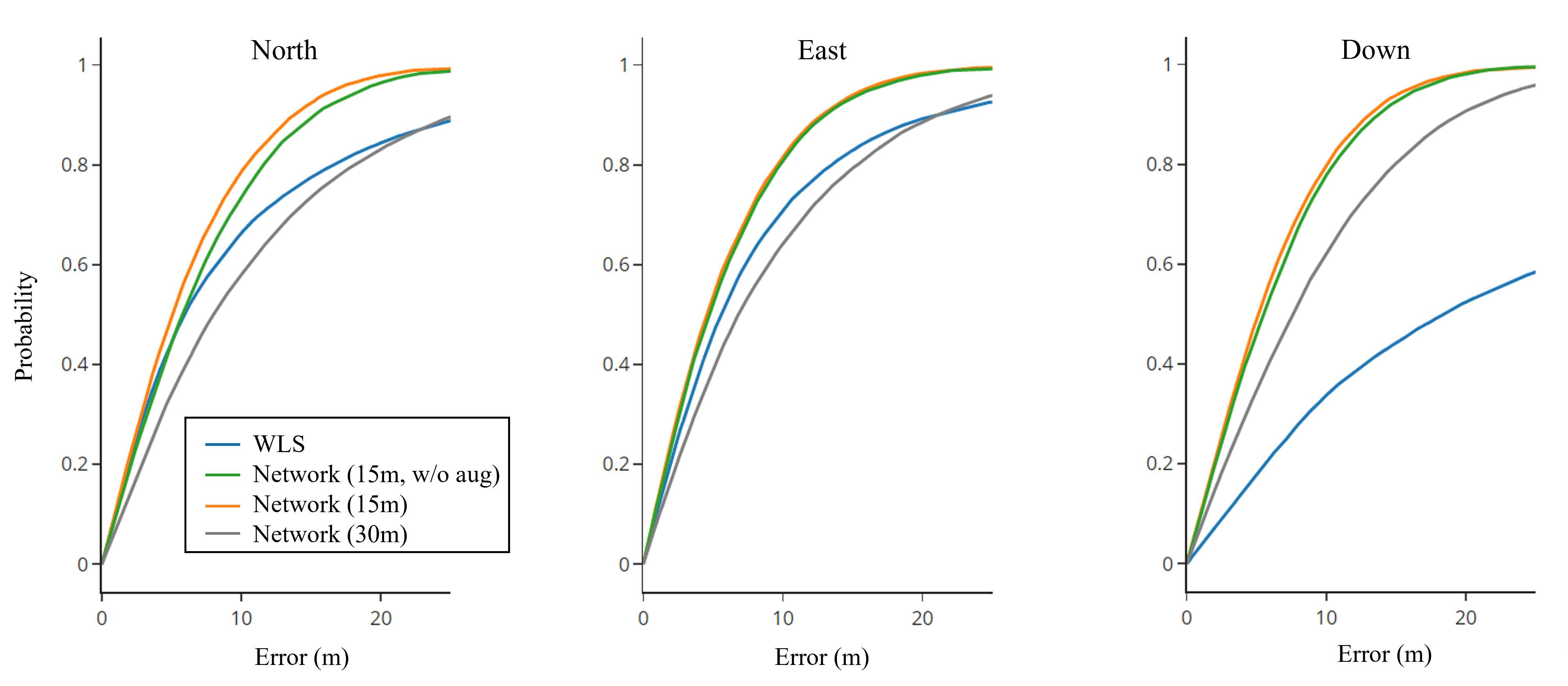}
    \caption{\add{Cumulative distribution function (CDF) of the absolute error, \newedit{on the test data,} along the local North, East and Down directions for our approach with $\eta=15\si{m}$, our approach without data augmentation, our approach with $\eta=30\si{m}$ and the WLS baseline. We observe that our approach with $\eta=15\si{m}$ outperforms all other variations and overbounds their corresponding CDFs. Our approach without data augmentation overbounds the WLS baseline as well. Our approach with $\eta=30\si{m}$ is overbound by WLS for smaller values of error in the North and East directions but overbounds the WLS baseline in the Down direction.}}
    \label{fig:cdf_plots}
\end{figure}

\add{We evaluate the performance of our proposed DNN with NED corrections and data augmentation using $\eta=15\si{m}$ to our approach without augmentation, our approach with data augmentation using $\eta=30\si{m}$ and the WLS baseline.
This evaluation is performed on the entire testing dataset and our experiments show that our approach with $\eta=15\si{m}$ performs the best out of all variations, both in terms of MAE, listed in Table~\ref{tab:android_ablative} and CDF plots of the errors, shown in Fig.~\ref{fig:cdf_plots}.

\newedit{We also evaluated a network that predicted positions directly, instead of predicting corrections to an initial position.
However, such an approach showed MAE in the order of $\si{10^3}{m}$ along all directions and was not investigated further or compared to other methods.}

Of the three variations of our method\newedit{, that we evaluated}, turning off the data augmentation has the least negative impact on the performance of the neural network. 
This difference is particularly noticeable in the North direction where the CDF curve deviations from the best case and an additional mean error of approximately $0.8\si{m}$ is observed.
The differences along the East and Down directions are not as evident, with an additional mean error of $0.15\si{m}$ to $0.25\si{m}$ and virtually indistinguishable CDF curves.

Similar to our observations from the simulated data, increasing the initialization range $\eta$ increases the MAE and causes a perceptible drop in the CDF curve for same error values.

Performance of the WLS baseline is poorer than both networks initialized with $\eta=15\si{m}$ in all three directions. 
However, the WLS baseline outperforms the network initialized with $\eta=30\si{m}$ in the North and East directions while still performing poorly in the down direction.}

\begin{figure}[t]
    \centering
    \includegraphics[width=\textwidth]{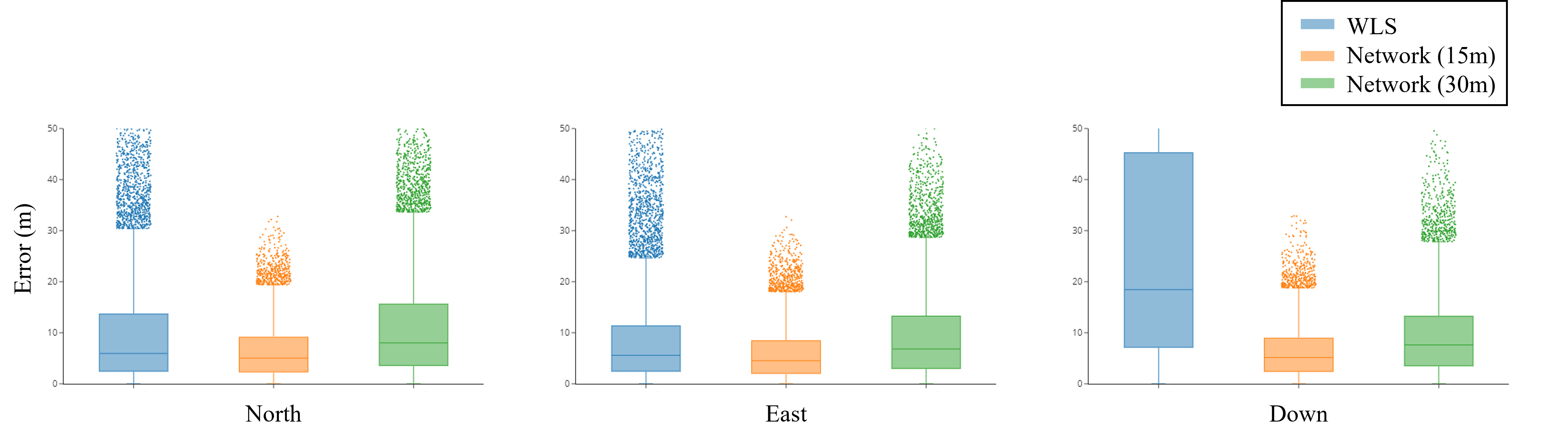}
    \caption{Localization error on the Android Raw GNSS Measurements Dataset~\citep{fu_android_2020} in the north, east and down directions respectively for WLS and our approach (initialization range $\eta=\siunit{15}{m}$ and $\siunit{30}{m}$). \add{The shaded regions cover the first quantile $Q1$ to the third quantile $Q3$, with the horizontal line representing the median. The vertical lines extend to a distance of $1.5\cdot \lvert Q3 - Q1 \rvert$ from $Q3$ and $Q1$ respectively. Points plotted beyond the vertical lines are samples at values greater than $Q3 +1.5\cdot \lvert Q3 - Q1 \rvert$ and lesser than $Q1 - 1.5\cdot \lvert Q3 - Q1 \rvert$.} \edit{Our approach with} $\eta=\siunit{15}{m}$ shows smaller localization error than WLS in all directions, while our approach with $\eta=\siunit{30}{m}$ shows smaller localization error than WLS in the down direction.}
    \label{fig:results_realdata}
\end{figure}

\edit{This difference is further evidenced by a comparison of the error quantiles between our approach with $\eta=15\si{m}$, our approach with $\eta=30\si{m}$ and the WLS baseline, as shown in Fig.~\ref{fig:results_realdata}.
Our approach with $\eta=15\si{m}$ outperforms the WLS baseline in all directions.
However, with $\eta=30\si{m}$, our approach is only able to outperform WLS in the down direction.}
\add{Similar to the simulated data, there is a strong correlation between the accuracy and the largest magnitude of the initial error, which is currently a limitation of this proposed work.}

\add{Fig.~\ref{fig:results_realdata} also demonstrates that the network learns the largest magnitude of error in the training dataset and bounds the estimated position correction using this information.
This also results in the improved performance of networks with smaller initialization ranges $\eta$ which provide corrections with correspondingly smaller magnitudes.}
\add{The network's initial guess is always within a certain range of the ground truth, because of which the network's final estimate is also relatively closer to the ground truth solution.
This results in our approach's superior performance on data not seen during training, like the urban case visualized in Fig.~\ref{fig:visual_realdata}.
The semi-urban case visualized in Fig.~\ref{fig:visual_realdata} is similar to data encountered by the network during training and consequently, the network's performs as expected in that case.}
\begin{table}[b]
    \centering
    \caption{\add{Mean absolute positioning error along the north, east and down directions in the estimate of the WLS baseline and variations of our approach (NED corrections + $\eta=30\si{m}$, NED corrections + $\eta=30\si{m}$ without data augmentation, and NED corrections + $\eta=15\si{m}$). We observe that smaller initialization range results in smaller position estimate errors, data augmentation improves performance on the testing dataset and that final positioning errors are significantly less than those of WLS estimates in the Down direction for all cases.}}
    \begin{tabular}{llll}
    \toprule
    \textbf{Scenario}       & \textbf{North (m)} & \textbf{East (m)} & \textbf{Down (m)} \\ \midrule
    WLS baseline            & $11.6 \pm 51.9$            & $9.7 \pm 38.7$            & $36.4 \pm 265.9$            \\
    Our approach with $\eta=30\si{m}$       & $11.1 \pm 10.2$           & $9.3 \pm 8.5$            & $9.3 \pm 7.5$           \\
    Our approach without data augmentation & $7.1 \pm 5.7$           & $6.0 \pm 5.1$            & $6.6 \pm 5.1$ \\ 
    \textbf{Our approach with} $\boldsymbol{\eta}\mathbf{=15\si{\textbf{m}}}$ & $\mathbf{6.4 \pm 5.2}$           & $\mathbf{5.9 \pm 5.0}$            & $\mathbf{6.2 \pm 4.9}$ \\ \bottomrule \\
    \end{tabular}
    \label{tab:android_ablative}
\end{table}

%% file: sections/06_conclusion.tex
\section{Conclusion}\label{sec:conclusion}
In this work, we proposed an approach to use a deep neural network (DNN) with GNSS measurements to provide a position estimate.
Our proposed approach is the first, to our knowledge, that works with GNSS measurements to provide outputs in the position domain.

To obtain a position estimate, we convert the traditional position estimation problem to that of estimating position corrections to an initial position guess using a DNN.
Our proposed approach addresses the challenge of set-based GNSS inputs, that vary in number and order, by utilizing the set transformer in the DNN architecture. 
We proposed using pseudorange residuals and LOS vectors from the initial position guess as inputs and NED position corrections as outputs to the DNN.
This particular choice of inputs and outputs improves the numerical conditioning of the DNN and provides \edit{a natural method to extend our approach to other} global \edit{regions}.
Additionally, to reduce overfitting on training data and incentivize the DNN to learn a functional map between the measurements and position corrections, we developed a geometry-based data augmentation method.

We validated our proposed approach on both simulated and real-world data. 
Experiments performed on the simulated data showed that the position corrections provided by the DNN reduced the \newedit{mean} absolute localization error in each of the North, East and Down directions from the error in the initial position guess, indicating that the DNN effectively learns to solve the positioning problem.
Experiments on real-world data demonstrated that the performance of the DNN is sensitive to the error present in the initial position guess.
Comparison of the absolute localization error to a weighted least squares (WLS) baseline showed that our approach outperforms WLS along the vertical direction when initialized with position errors within $\siunit{15}{m}$ as well as $\siunit{30}{m}$.
\add{Our experimentation also validates that our data augmentation technique improves the network's performance, when compared to a similar network without data augmentation.}

\add{This work validates that using DNNs for GNSS-based localization is a promising and interesting area of research.
Our current approach is a snapshot method limited to using simple features. 
Additionally, both our training and testing
datasets are entirely from the San Francisco Bay Area, which does not provide geographical diversity. 
In the future, we plan to validate our proposed method on diverse testing datasets collected from locations around the globe.
We also plan to extend our approach to sequential position estimation while considering additional measurements such as signal-to-noise-ratio and Doppler. 
Furthermore, we are considering performing a more detailed parametric study to investigate the effect of hyperparameter values, the use of additional regularization methods and an iterative positioning correction approach similar to CMR Net~\citep{cattaneo_cmrnet_2019}.} 
\add{Our proposed work is also limited by its reliance on close initial guesses and the sensitivity to initialization ranges, which we will also address in future work.}


%% file: sections/07_acknowledgements.tex
\section*{acknowledgements}
Some of the computing for this project was performed on the Sherlock cluster. We would like to thank Stanford University and the Stanford Research Computing Center for providing computational resources and support that contributed to this research. 
